\documentclass{article}

\PassOptionsToPackage{numbers, compress}{natbib}

\usepackage[preprint]{neurips_2023}

\usepackage[utf8]{inputenc} 
\usepackage[T1]{fontenc}    
\usepackage{url}            
\usepackage{booktabs}       
\usepackage{amsfonts}       
\usepackage{nicefrac}       
\usepackage{microtype}      
\usepackage{xcolor}         
\usepackage{graphicx}
\usepackage{subfig}
\usepackage{multirow}
\usepackage[pagebackref=true,breaklinks=true,colorlinks,urlcolor=black,bookmarks=false]{hyperref}

\def \robotlogo {\raisebox{-0.1\height}{\includegraphics[height=0.95\baselineskip]{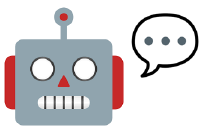}}}

\newcommand{\eg}{\textit{e}.\textit{g}.}

\newcommand\blfootnote[1]{%
	\begingroup
	\renewcommand\thefootnote{}\footnote{#1}%
	\addtocounter{footnote}{-1}%
	\endgroup
}

\title{\robotlogo{} \textcolor{cyan}{I}ntern\textcolor{cyan}{GPT}: Solving Vision-Centric Tasks \\ by Interacting with ChatGPT Beyond Language}

\author{\small
	\hspace{-0.38cm}\textbf{Zhaoyang Liu$^{*1}$, Yinan He$^{*1}$, Wenhai Wang$^{*\dagger 1}$, Weiyun Wang$^{*1}$, Yi Wang$^{*1}$, Shoufa Chen$^{*2,1}$} \\
	\small\hspace{-0.38cm}\textbf{Qinglong Zhang$^{*1}$, Zeqiang Lai$^{*3,1}$, Yang Yang$^1$, Qingyun Li$^1$, Jiashuo Yu$^1$, Kunchang Li$^{4,1}$, Zhe Chen$^{5,1}$, } \\ 
	\small\hspace{-0.38cm}\textbf{Xue Yang$^1$, Xizhou Zhu$^{6,1}$, Yali Wang$^{4,1}$, Limin Wang$^{5,1}$, Ping Luo$^{2,1}$, Jifeng Dai$^{7,1}$, Yu Qiao$^1$} \\
	\small\hspace{-0.38cm}$^1$OpenGVLab, Shanghai AI Laboratory \quad $^2$The University of Hong Kong \quad $^3$Beijing Institute of Technology\\
	\small\hspace{-0.38cm}$^4$Shenzhen Institutes of Advanced Technology, Chinese Academy of Sciences \\
	\small\hspace{-0.38cm}$^5$Nanjing University \quad $^6$SenseTime Research \quad $^7$Tsinghua University \\
	{\url{https://github.com/OpenGVLab/InternGPT}} \\
}

\begin{document}

	\maketitle
	
	\emph{We're going to use the best pointing device in the world. We're going to use a pointing device that we're all born with --- born with ten of them. We're going to use our fingers. We're going to touch this with our fingers.}
	
	\rightline{--- Steve Jobs}
	
	\begin{abstract}
		We present an interactive visual framework named InternGPT, or iGPT for short. 
		The framework integrates chatbots that have planning and reasoning capabilities, such as ChatGPT, with non-verbal instructions like pointing movements that enable users to directly manipulate images or videos on the screen. Pointing (including gestures, cursors, etc.) movements can provide more flexibility and precision in performing vision-centric tasks that require fine-grained control, editing, and generation of visual content. 
		The name InternGPT stands for \textbf{inter}action, \textbf{n}onverbal, and \textbf{chat}bots. 
		Different from existing interactive systems that rely on pure language, by incorporating pointing instructions, the proposed iGPT significantly improves the efficiency of communication between users and chatbots, as well as the accuracy of chatbots in vision-centric tasks, especially in complicated visual scenarios where the number of objects is greater than 2. Additionally, in iGPT, an auxiliary control mechanism is used to improve the control capability of LLM, and a large vision-language model termed Husky is fine-tuned for high-quality multi-modal dialogue (impressing ChatGPT-3.5-turbo with 93.89\% GPT-4 Quality).
		We hope this work can spark new ideas and directions for future interactive visual systems.

	\end{abstract}

	\section{Introduction}
	
	Vision-centric tasks aim to enable computers to understand what they see from the world and react accordingly. In the past, these tasks were solved one by one using specific vision foundation models~(VFMs) that were pre-defined and trained for specific visual concepts (\eg, classes, masks, etc.). However, this approach is limited by the availability and quality of labeled data and the diversity of visual scenarios. Recently, the blossom of large language models~(LLMs) such as ChatGPT~\cite{openai2022chatgpt}, GPT-4~\cite{openai2023gpt4}, and LLaMA~\cite{touvron2023llama} has opened up new possibilities for solving vision-centric tasks. This paradigm leverages LLMs to learn unified real-world concepts and make decisions or plans with vision foundation models~(VFM). This approach is user-friendly, requiring almost no domain knowledge for daily or professional tasks via dialogues.
	It has led to the development of various killer applications, \eg, Visual ChatGPT~\cite{wu2023visual}, MM-REACT~\cite{Yang2023MMREACTPC}, HuggingGPT~\cite{shen2023hugginggpt}, etc.\blfootnote{* Equal contribution.\ \ \ \ $\dagger$ Project lead}
	
	Although words are convenient for defining tasks and describing objects, actions, scenes, events, etc, and language-based instructions powered by LLMs allow us to enjoy the capabilities of AI systems, current interactive systems~\cite{wu2023visual, Yang2023MMREACTPC, shen2023hugginggpt} have limitations in connecting vision and language models. They rely mainly on text instructions to interact with visual instances. This becomes highly inefficient when dealing with complicated visual scenarios involving multiple instances since we need to describe the desired instance in length by giving details to discriminate it from others.
	
	\begin{figure}[t]
		\centering
		\includegraphics[width=\linewidth]{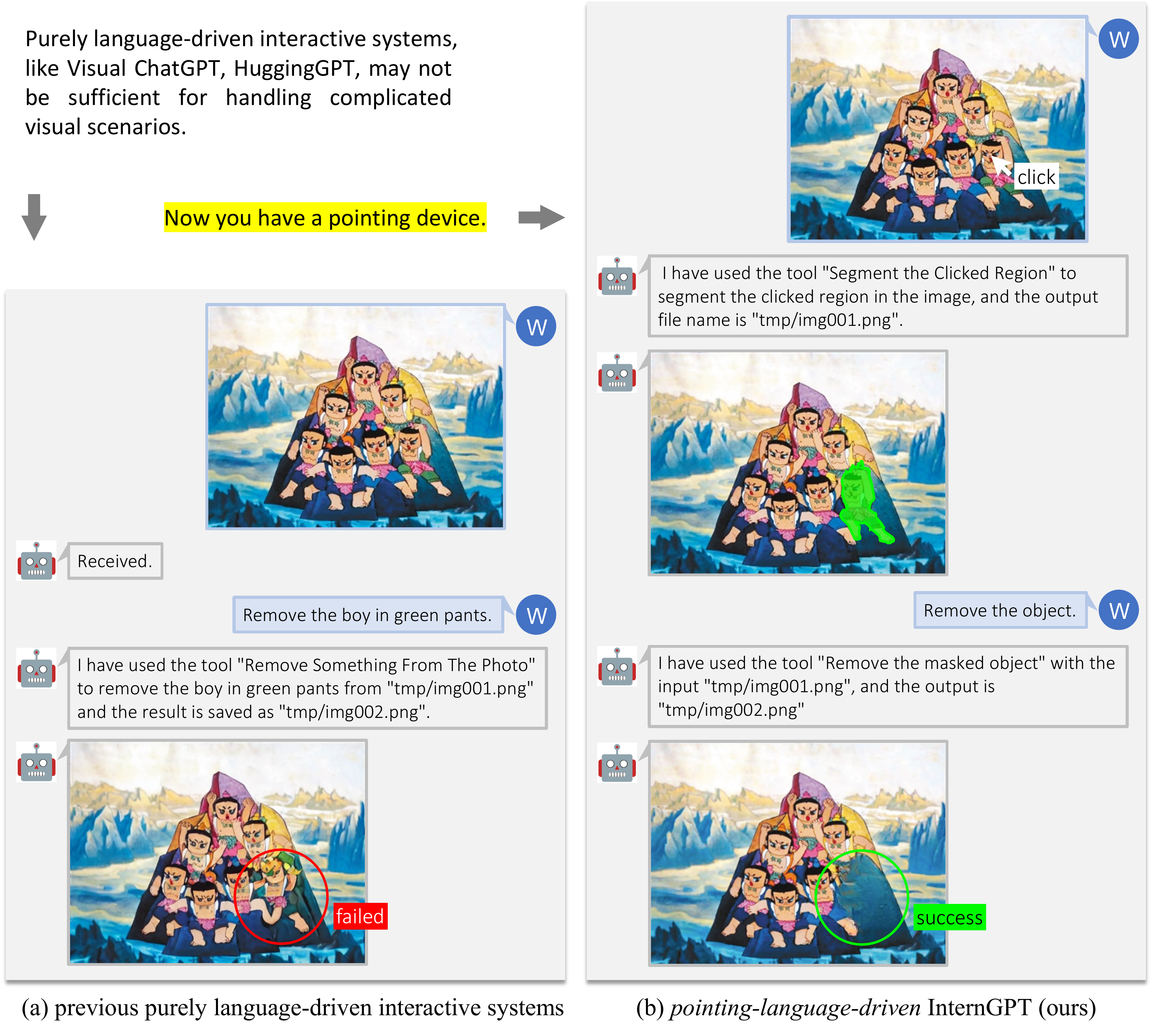}
		\caption{
			\textbf{Advantage of our \emph{pointing-language-driven} interactive system.} 
		}\label{fig:moti}
	\end{figure}
	
	Pointing movements such as gestures and cursors are the most common form of non-verbal instructions for communication, showing accuracy in pointing and brevity in operations. Modern operating systems rely heavily on pointing devices such as touchscreens and mice for navigation and interaction. To overcome the aforementioned limitations in current interactive systems, we need to explore new methods to integrate visual perception and language understanding, including both verbal and non-verbal instructions. This will enable efficient and accurate interactions in complicated visual environments.

\begin{figure}[!th]
		\centering
		\includegraphics[width=\linewidth]{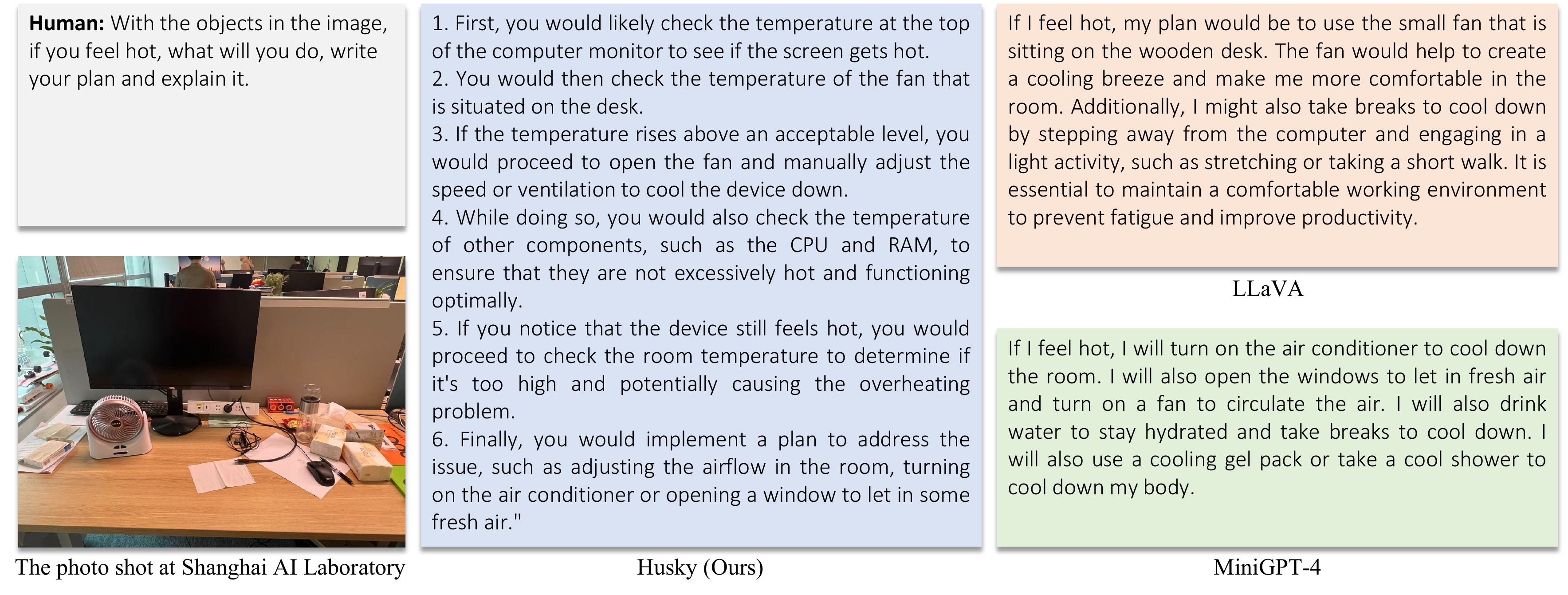}
		\caption{
			\textbf{Comparsion results of Husky, MiniGPT-4 and LLaVA.} 
		}\label{fig:husky_8}
	\end{figure}

	Inspired by this, we present an advanced interactive visual framework, termed InternGPT~(iGPT), which combines the advantages of pointing and language instructions to perform complex vision-centric tasks. It consists of three main components: a perception unit that handles pointing instructions on images or videos, an LLM controller with an auxiliary control mechanism that can accurately parse language instructions, an open-world toolkit that integrates various online models from HuggingFace, private models trained by users, as well as other applications (\eg, calculator, search engine). InternGPT gives equal importance to pointing and language instructions and uses the perception unit and the LLM controller to coordinate and execute the applications in the toolkit to accomplish complicated vision-centric tasks. In addition, for high-quality multi-modal dialogue including image captioning and visual question answering, we provide a large-scale visual language model named Husky. Husky is quite efficient and effective, which impresses ChatGPT-3.5-turbo with 93.89 \% GPT-4 Quality with only 7B parameters. Comparison between Husky with MiniGPT-4~\cite{zhu2023minigpt} and LLaVA~\cite{liu2023llava} are shown in Figure~\ref{fig:husky_8}, and we see that Husky works better in complex reasoning scenarios at a photo casually shot at Shanghai AI Laboratory.
	
	Compared with previous pure-language interactive frameworks, iGPT excels in achieving higher accuracy and efficiency in complicated  visual tasks including accurate interactive image or video editing, understanding, question-answering, visual content creation, etc. As evidenced in Figure~\ref{fig:moti}, our system can successfully carry out intricate interactive tasks while pure-language systems fail.
	Moreover, our user surveys have revealed that combining pointing instructions with language instructions can enhance work efficiency significantly, particularly in complicated scenarios that involve many objects (object number greater than 2). We aim to establish this work as an open baseline for visual interactive systems and will continue updating it with the capability of our VFMs~(\eg, InternImage~\cite{wang2022internimage} and InternVideo~\cite{wang2022internvideo}) and the contribution/pull request from the community to achieve even better results.

	\section{Related Work}
	\textbf{Large language model.}
	Recent LLMs~\cite{brown2020gpt3,openai2023gpt4,touvron2023llama,zeng2022glm} have demonstrated a range of significant abilities, including language generation, in-context learning, world knowledge, and reasoning. The presence of these capabilities enables LLMs to perform complex tasks based on user instructions and prompts in a zero-shot manner.
	GPT-3~\cite{brown2020gpt3}, the first language model with over 100 billion parameters, has achieved impressive zero-shot performance on various benchmarks. However, it does not consistently outperform smaller models, such as T5~\cite{raffel2020t5}, on some tasks.
	InstructGPT models~\cite{ouyang2022instruct-tuning}, which are finetuned on a dataset consisting of prompts with the corresponding human-annotated desired behavior, can be aligned with users, generate outputs that are preferred over those from GPT-3 and show improvements in truthfulness and reductions. Instruction-tuned models have also demonstrated a remarkable ability to generalize zero-shot to new tasks. Thus, instruction-tuning~\cite{longpre2023flan, wang2022super, iyer2022opt-iml, chung2022scaling} is considered key to eliciting the abilities of LLMs~\cite{fu2022gptroadmap}.
	In addition to GPT model family \cite{radford2018gpt1,radford2019gpt2,brown2020gpt3,openai2023gpt4}, several other LLMs exist, including OPT~\cite{zhang2022opt}, LLaMA~\cite{touvron2023llama}, MOSS~\cite{moss} and GLM~\cite{zeng2022glm}. These models also achieve high performance and are open-sourced, providing valuable experience in training large models and serving as a base for further fine-tuning for different purposes. For example, Alpaca~\cite{wang2022self_instruct} proposes a self-instruct framework to instruction tune the LLaMA model family without relying heavily on human-written instruction data.
	Another active research area on LLMs is chain-of-thought prompting (CoT)~\cite{wei2022chain,kojima2022large,gao2022pal,wang2022self}. CoT prompts models to solve problems step by step, greatly improving their reasoning ability of LLMs and making it possible to utilize LLMs for task splitting. As a result, LLMs can be combined with a variety of APIs \cite{li2023apibank,nakano2021webgpt} and models \cite{shen2023hugginggpt,wu2023visual} trained for different modalities and serve as a controller to schedule them.
	This method liberates LLMs from pure language instructions and paves the way for a multi-modal interactive system.

	\textbf{Perception model.}
	The emergence of the AlexNet~\cite{krizhevsky2017imagenet} can be considered as the beginning of the development history of deep convolutional neural networks (CNN). 
	Drawing on the success of AlexNet, many CNN with deeper networks, more parameters, and better performance have been proposed and successfully applied to computer vision. However, blindly deepening the network and increasing parameters will not improve the performance without limit but will cause overfitting and increase the cost of experiments. In 2004, GoogleNet~\cite{szegedy2015going} overcame the above issues by processing images at multiple scales thanks to the proposed Inception mechanism, which combines convolution operations and pooling operations of different core sizes. In the same year, the concise VGG~\cite{simonyan2015very} only used (3$\times$3) convolution and (2$\times$2) pooling to win the second place in classification and the first place in object detection. The ResNet~\cite{he2016deep} has extended the number of layers of the network to an unprecedented scale and solved the issue of deep network degradation, finally achieving 3.57\% error on the ImageNet~\cite{deng2009imagenet} test set. The above models have achieved the dominant position of CNN in the visual field, and it was not until the birth of Vision Transformer~(ViT)~\cite{dosovitskiy2021image} that this pattern was changed. Benefiting from the powerful Transformer~\cite{vaswani2017attention} structure, many more advanced transformer-based vision models, \eg, PVT~\cite{wang2021pyramid,wang2022pvt}, Swin Transformer~\cite{liu2021swin}, etc., have been proposed. At the same time, CNN has also been revived (\eg, ConvNeXt~\cite{liu2022convnet}, InternImage~\cite{wang2022internimage}), and some hybrid methods (\eg, ConViT~\cite{d2021convit}, CeiT~\cite{yuan2021incorporating}, and CoAtNet~\cite{dai2021coatnet}) of CNN and Transformer have appeared to make full use of the advantages of both.
	The benign competition of the visual foundation model has also greatly promoted the development of other visual recognition tasks, such as object detection~\cite{ren2015faster, lin2017focal, carion2020end, zhu2021deformable, yang2019scrdet,yang2021r3det,yang2021rethinking,yu2023h2rbox}, segmentation~\cite{cheng2021per, jain2022oneformer, li2022panoptic, chen2022vision, ji2023ddp, girdhar2023imagebind}, video understanding~\cite{Wang2016TemporalSN, Lin2018TSMTS, Feichtenhofer2018SlowFastNF, Bertasius2021IsSA, Arnab2021ViViTAV, Li2022UniFormerUT, videomae, chen2022internvideo, wang2023videomae, li2022uniformerv2, li2023unmasked, yang2023basictad,liu2021tam}, etc.
	These models have a strong ability to perceive the physical world and can serve as the sensory organs for LLMs.

	\textbf{LLM-based interactive system.}
	The success of LLMs~\cite{brown2020gpt3, openai2023gpt4} has led to the development of AI systems that integrate perception models and LLMs for multimodal reasoning and action. 
	One such system is Visual ChatGPT~\cite{wu2023visual}, which connects ChatGPT with visual foundation models to generate and edit images during chatting. 
	Another paradigm is MM-REACT~\cite{Yang2023MMREACTPC}, which integrates ChatGPT with a pool of vision experts for multimodal reasoning and action. 
	HuggingGPT~\cite{shen2023hugginggpt} utilizes numerous sophisticated AI tasks in different modalities and domains from HuggingFace's abundant AI models to achieve impressive results. 
	TaskMatrix.AI~\cite{liang2023taskmatrix} connects foundation models with millions of APIs from other AI models and systems to perform diversified tasks in both digital and physical domains.
	However, these systems are limited by their dependence on pure language instructions, which can hinder effective communication and task performance, just like a console cannot achieve what GUI can.
	By incorporating pointing instructions, the proposed iGPT significantly improves the efficiency~of~communication between users and chatbots, as well as the accuracy of chatbots in vision-centric tasks.

	\section{InternGPT}

	\begin{figure}[t]
		\centering
		\includegraphics[width=\linewidth]{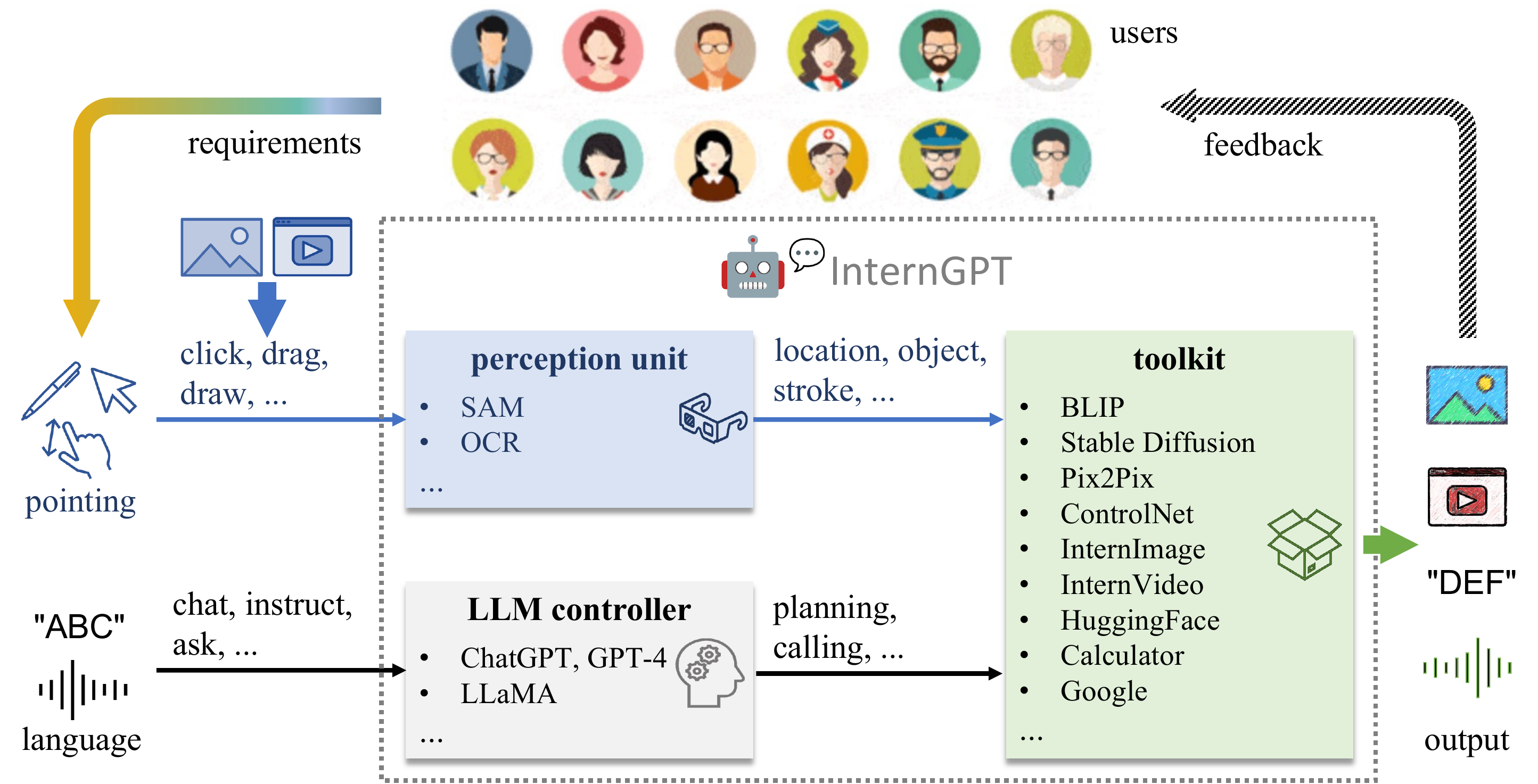}
		\caption{\textbf{Overall architecture of InternGPT}. It has three main components: perception unit, LLM controller, and open-world toolkit.}
		\label{fig:arch}
	\end{figure}
	
	InternGPT (iGPT) aims to provide an intuitive, user-friendly, and efficient way of human-computer interaction (HCI) in AI that leverages the integration of a large language model (LLM), pointing devices, and computer vision algorithms to perform vision-centric tasks.
	As illustrated in Figure~\ref{fig:arch}, iGPT consists of three main components: (1) a perception unit that interprets the user’s pointing gestures on images and videos, enabling precise object selection and identification; (2) an LLM controller that processes the user’s language commands, facilitating natural communication and (3) an open-world toolkit that integrates various off-the-shelf models/applications to offer a versatile platform for different tasks.
	
	iGPT's design allows it to operate effectively at multiple levels, catering to diverse needs:
	
	\textbf{Level 1: basic interaction.}
	An intuitive way to use iGPT is to give it simple commands that trigger pre-defined tasks, such as ``\texttt{caption this photo}''. iGPT then calls the appropriate model, such as BLIP~\cite{li2022blip}, to produce the desired results.
	At this level, iGPT acts as a front-end of traditional single-task foundation models, without requiring complicated interactive logic, such as chain of thought, contextual reasoning, etc.
	
	\textbf{Level 2: Language-guided interaction.}
	Real-world tasks often involve more complex and diverse demands than traditional pre-defined tasks at level 1. To accomplish these tasks, language instructions with clear specifications or multiple dialogue turns are needed. For example, the command ``\texttt{remove the black dog near the table in the image}'' requires open-vocabulary detection models to identify the target object based on the language description, and then apply the erase model to remove it from the image.
	At this level, iGPT is an assistant that communicates with users to resolve any ambiguity using natural language.
	
	\textbf{Level 3: pointing-language enhanced interaction.}
	When tasks require more precise specifications that language instructions alone cannot provide, pointing-language instructions become essential. For example, editing/recognizing/OCRing a particular part of an image can be difficult to describe with words. Nonverbal cues, such as gestures or cursor movements, help select, move, or draw objects in the image.

	Compared to existing systems like ChatGPT and Visual ChatGPT, \emph{iGPT represents a significant advancement in user-centric interaction by combining pointing and language instructions to accomplish complex vision-centric tasks.} Practical use cases include image editing, object manipulation, video annotation, and more, demonstrating its potential impact across various academic and industrial fields. We detail iGPT's design in the following.

	\subsection{Interacting with User}
	
	As shown in Figure \ref{fig:arch}, iGPT's framework accomplishes its tasks through continuous interaction with users.
	Each time the user provides their requirements via pointing and language instructions, the perception unit analyzes the pointing instructions to identify the target or the content created.
	Simultaneously, the LLM controller interprets the user's language instructions, breaking down the task into smaller subtasks and selecting the appropriate tool.
	
	These components work in tandem, delivering accurate and efficient results to users.
	Similar to previous methods~\cite{wu2023visual}, our system keeps a record of the conversation history, enabling users to revisit prior tasks and results.
	This capability ensures continuous improvement and better performance~over~time.

	\subsection{Perception Unit}

	Built on community open-source projects like SAM~\cite{kirillov2023segment} and OCR~\cite{easyocr}, iGPT's perception unit parses pointing instructions and performs various operations, such as pick, drag, and draw. It uses click, stroke, drag, and draw gestures for object selection, movement, and content creation. Specifically, the drag gesture is used to move objects to different positions, while the draw gesture is used to create or complete shapes, aided by image generation technology.
	
	The current system uses simple logic to interpret pointing instructions. After the pointing gesture is finished, we handle it in three ways: (1) For typical objects, SAM detects the semantic region to enable selection.
	(2) For scene text, OCR technology extracts pointed text content.
	(3) For generation tasks, gestures are stored as strokes and fed into AIGC tools.
	
	The perception unit's versatility and proficiency in executing various operations based on user input are crucial to our system's success. Whether the user needs to select an object, move it, or create new content, the perception unit makes it possible. By combining cutting-edge AI solutions with user-friendly interfaces, we ensure a seamless and engaging user experience.

	\subsection{LLM Controller}
	
	Similar to previous interactive systems~\cite{wu2023visual,Yang2023MMREACTPC,shen2023hugginggpt}, iGPT manages complex tasks based on large language models (LLMs). It autonomously parses user language requests, decomposes them into multiple tasks, and plans the task order and dependency based on LLM knowledge. The LLM controller allocates the parsed tasks to the corresponding APIs based on the model descriptions. Through continuous analysis of historical data and user interactions, the system improves task allocation and execution for efficient management.
	
	\textbf{Accurate task execution with the auxiliary control mechanism.} Even top language models like ChatGPT and GPT-4 struggle with invoking APIs, especially when parsing and passing arguments. To tackle this issue, iGPT employs auxiliary control when LLM fails to act as a controller. It works as follows: (1) Parse verbs and nouns to identify the API before execution. (2) Extract relevant arguments from prior dialogues based on the API. For example, in the case of ``remove the masked object'', query the LLM: ``\texttt{What’s the image\_path and mask\_path of the `remove the masked object' API?}'' This retrieves the ``\texttt{image\_path}'' and ``\texttt{mask\_path}'' from the chat history. (3) Check argument validity and apply rule-based corrections if necessary. (4) Invoke the API with the identified arguments.
	This mechanism refines abstract instructions into specific commands, resulting in more accurate task execution.
	
	\textbf{Speech transcription.} Additionally, iGPT also offers speech transcription features, such as whisper \cite{radford2022robust} and bark~\cite{bark}, enabling users to communicate without typing.
	
	\subsection{Toolkit}
	
	iGPT's toolkit is called upon by the system. Different from previous works \cite{wu2023visual, Yang2023MMREACTPC, shen2023hugginggpt}, it supports input from pointing devices in addition to standard API descriptions, inputs, and outputs.
	Here, we show four representative examples of API descriptions as follows:
	
	\textbf{Example 1: remove the masked object.} \textbf{Input:} original image, mask (pick); \textbf{Output:} result image; \textbf{Prompt:} ``\texttt{useful when you want to remove an object by masking the region in the image, like: remove the object by the masked region. The input to this tool should be a comma-separated string of two, representing the image\_path and mask\_path. The input to this tool should be a string, representing the image\_path}''.
	
	\textbf{Example 2: question the masked object.} \textbf{Input:} original image, mask (pick); \textbf{Output:} result text; \textbf{Prompt:} ``\texttt{useful when you need an answer for a question based on a masked image. like: what is the background color in the masked region, how many cats are in this masked figure, what is in this masked figure. The input to this tool should be a comma-separated string of two, representing the image\_path and the question}''.
	
	\textbf{Example 3: conditional image generation.} \textbf{Input:} stroke draft (drag \& draw); \textbf{Output:} result image; \textbf{Prompt:} ``\texttt{useful when you want to replace an object by clicking in the image with another object or something. like: replace the masked object with a new object or something. The input to this tool should be a comma-separated string of three, representing the image\_path and the mask\_path and the prompt}''. 
	
	\textbf{Example 4: video highlight interpretation.} \textbf{Input:} original video, mask at timestamp $t$ (pick); \textbf{Output:} result video; \textbf{Prompt:} ``\texttt{useful when you want to generate a video with TikTok style based on prompt, like: cut this video to a TikTok video based on a prompt. The input to this tool should be a comma-separated string of two, representing the video\_path and prompt.}''

	A comprehensive summary of the toolkit's APIs is provided in Table \ref{tab:apis}, covering applications including vision, vision-language foundation models, as well as other applications such as calculators and search engines. This diverse range of APIs enables users to harness advanced techniques for various tasks and achieve their goals more efficiently.
	
	Specifically, for the large vision-language model named Husky in Table \ref{tab:apis}, we follow the approach of BLIP-2~\cite{li2023blip}, and replace the language model with LLaMA-7B,  which is trained on 52k English instruction-following data generated by GPT-4~\cite{openai2023gpt4}. To handle multi-modal tasks without compromising the text-only generation capabilities, we strictly adhere to the input format used in instruct-tuned LLaMA and carry out finetuning of Husky in three stages. As a result, Husky exhibits impressive capabilities, such as visual captioning, visual question-answering, complex reasoning, code generation similar to those of MiniGPT-4~\cite{zhu2023minigpt} and LLaVA~\cite{liu2023llava}.
	
	\begin{table}[ht]
		\centering
		\begin{tabular}{ l | c }
			\toprule
			Category &  API  \\
			\midrule
			\multirow{2}{*}{Vision}   & Stable Diffusion~\cite{rombach2021highresolution}, ControlNet~\cite{zhang2023adding}, InternImage~\cite{wang2022internimage}, InternVideo~\cite{wang2022internvideo}, \\ & SAM~\cite{kirillov2023segment}, DINOv2~\cite{oquab2023dinov2}, DragGAN~\cite{pan2023draggan}, ImageBind~\cite{girdhar2023imagebind}, etc. \\
			\midrule
			Vision-Language & BLIP-2~\cite{li2023blip}, Grounding DINO~\cite{ShilongLiu2023GroundingDM}, GLIP~\cite{li2021grounded}, VideoChat~\cite{2023videochat}, Husky, etc. \\
			\midrule
			Others & Calculator, Google, HuggingFace, etc. \\
			\bottomrule
		\end{tabular}
		\vspace{1em}
		\caption{\textbf{Representive APIs supported by InternGPT.}}
		\label{tab:apis}
	\end{table}
	
	\section{Experiment} 
	\subsection{User Studies}

	The iGPT framework enhances the communication effectiveness of the interactive system with the user by incorporating verbal and nonverbal instructions. To prove the effectiveness of our mixed-mode instructions, which combine language and pointing gestures, in comparison to purely linguistic instructions, we conduct a user study involving 10 human participants.  These participants interact with Visual ChatGPT~\cite{wu2023visual} and our iGPT through chat and provide their feedback. We present various findings from the user study as follows:
	
	\textbf{Efficiency.}
	We ask users to create an image-centric work by designing input instructions for Visual ChatGPT~\cite{wu2023visual} and iGPT, which involve removing and replacing objects. Users can refine their instructions if the results are unsatisfactory. We assume that a user has the patience to attempt up to ten times. Cases with over ten attempts are considered as failed. The number of attempts and prompt lengths required to achieve satisfactory results are presented in Table~\ref{tab:user_study1} and Table~\ref{tab:user_study2}, demonstrating that the iGPT is more efficient and user-friendly. 
	
	\textbf{Human preference.} 
	Users were asked to rank the interactive systems based on their user experience, focusing specifically on the results generated by these systems. Ten unbiased human evaluators were then responsible for assessing the quality of the outcomes. The evaluation results can be found in Table~\ref{tab:user_study1} and Table~\ref{tab:user_study2}. As can be observed, our iGPT stands out for its efficiency and user-friendliness, and gains a higher preferences.

	\newcommand{\imgitem}[1]{\multirow{6}{*}  {\includegraphics[height=0.11\linewidth]{#1}}}
	
	\begin{table}[ht]
		\small
		\tabcolsep=0.06cm
		\begin{tabular}{@{}l|ccccc@{}}
			\toprule
			& One-Object  &  Two-Object & Three-Object & Complex-I & Complex-II \\
			\midrule
			\multirow{6}{*}{Example} &  \imgitem{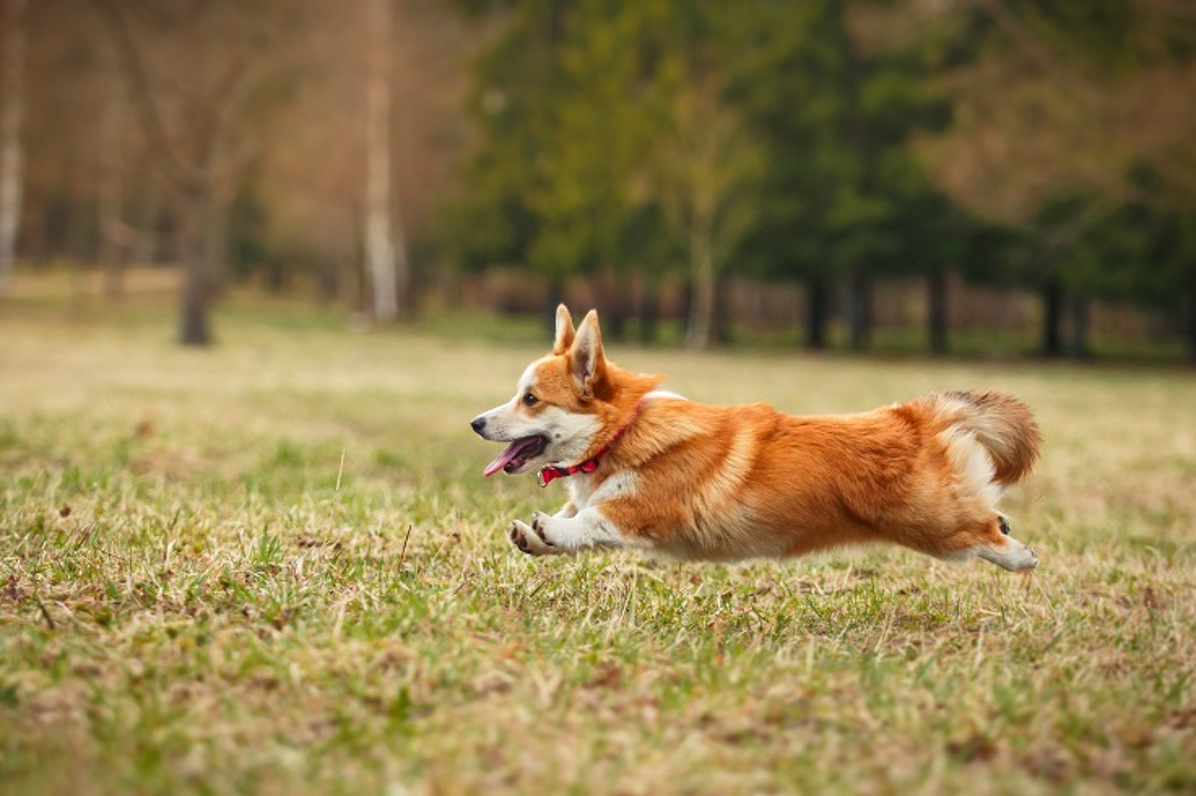}
			&  \imgitem{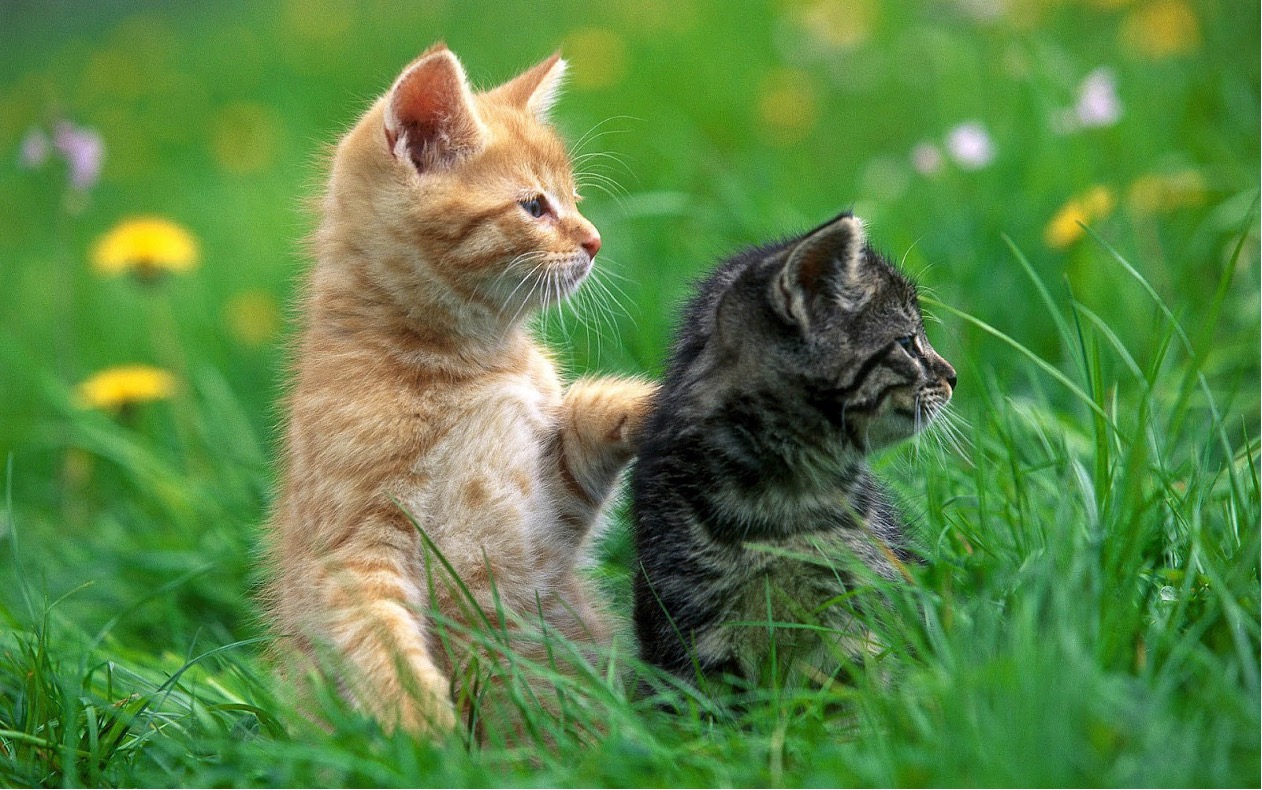}
			&  \imgitem{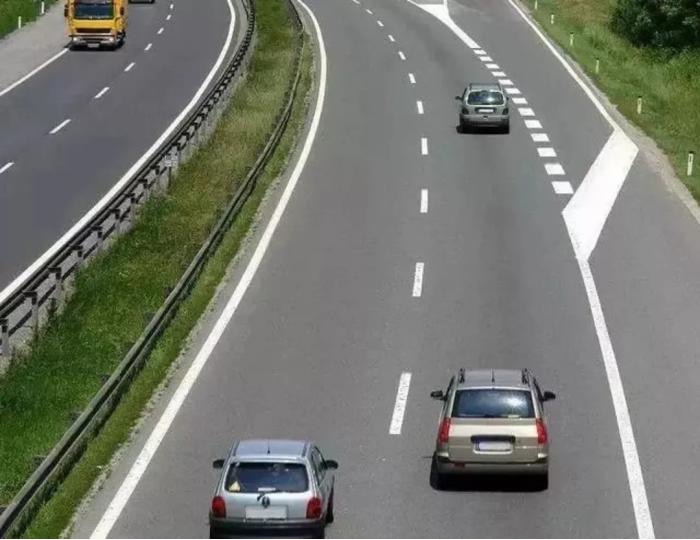}
			&  \imgitem{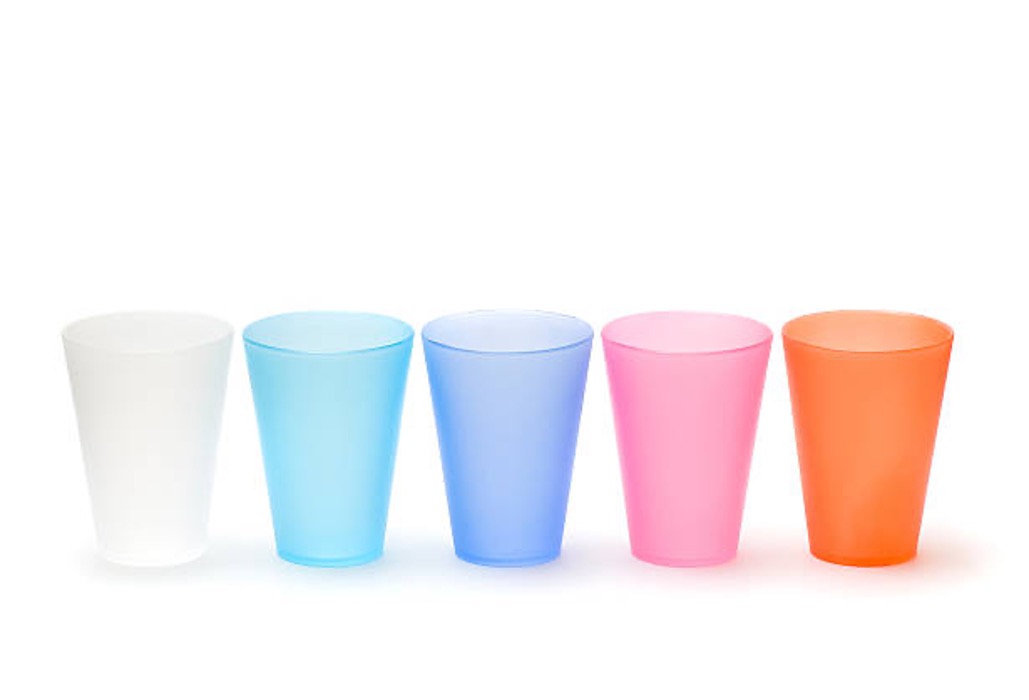}
			&  \imgitem{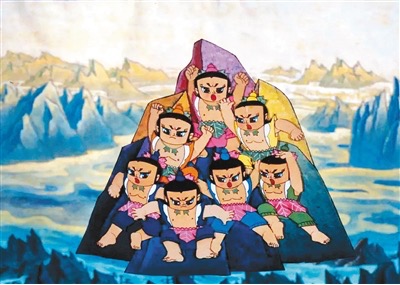} \\ 
			& & & & \\ 
			& & & & \\
			& & & & \\
			& & & & \\ 
			& & & & \\ \midrule
			& \multicolumn{5}{c}{\textbf{Visual ChatGPT}~\cite{wu2023visual}} \\
			Prompt length & 4.9 & 7.0 & 8.2 & 7.3 & 9.3 \\
			\# Try (<10) & 3.9 & 9.5 & 6.2 & 9.6 & 9.6 \\
			Score (1-5) & 3.2 & 1.5 & 2.6 & 0.3 & 0.6 \\
			\multirow{6}{*}{Example Result} &  \imgitem{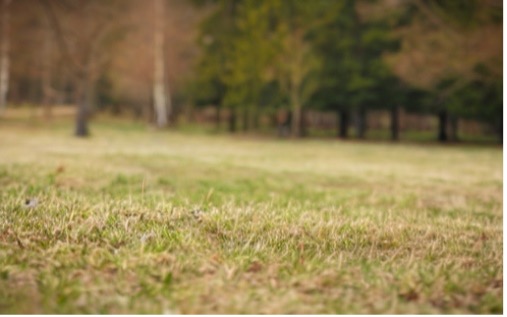}
			&  \imgitem{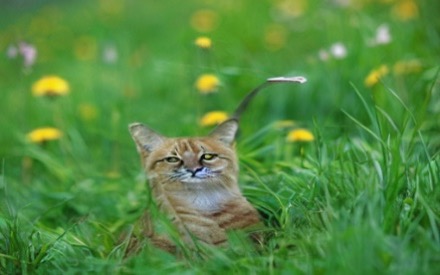}
			&  \imgitem{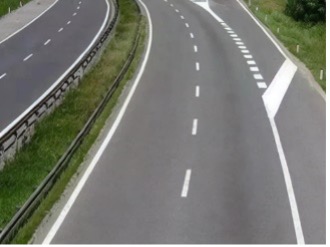}
			&  \imgitem{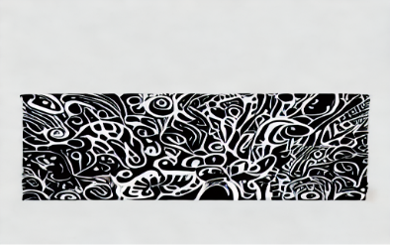}
			&  \imgitem{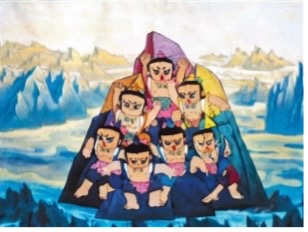} \\ 
			& & & & \\ 
			& & & & \\
			& & & & \\
			& & & & \\ 
			& & & & \\ \midrule
			& \multicolumn{5}{c}{\textbf{InternGPT (ours)}} \\
			Prompt length &  2.7 & 2.7 & 3.0 & 2.6 & 3.3 \\
			\# Try (<10) & 1.0 & 1.0 & 1.3 & 1.2 & 1.0 \\
			Score (1-5) & 4.9 & 3.6 & 3.2 & 4.5 & 3.4 \\
			\multirow{6}{*}{Example Result} &  \imgitem{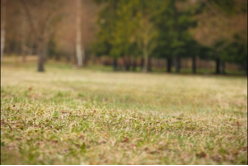}
			&  \imgitem{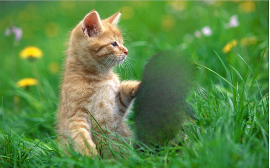}
			&  \imgitem{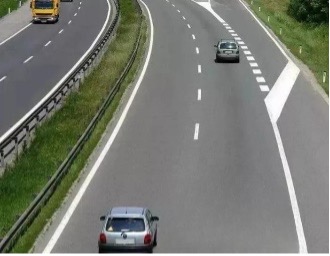}
			&  \imgitem{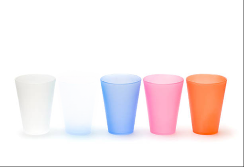}
			&  \imgitem{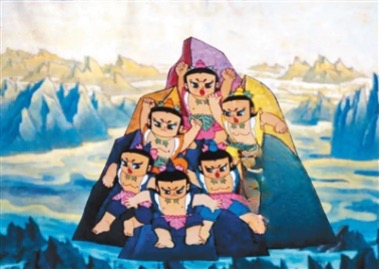} \\ 
			& & & & \\ 
			& & & & \\
			& & & & \\
			& & & & \\ 
			& & & & \\ \bottomrule
		\end{tabular}
		\vspace{1em}
		\caption{\textbf{User study for ``\texttt{remove something}''}. ``\# Try'' indicates the number of attempts to get a satisfactory result, not more than 10 times. ``-'' means trying more than 10 times and still getting bad results.}
		\label{tab:user_study1}
	\end{table}

	\begin{table}[ht]
		\small
		\tabcolsep=0.07cm
		\begin{tabular}{@{}l|ccccc@{}}
			\toprule
			Test Case  & Single-Object  &  Two-Object & Three-Object & Complex-I & Complex-II \\
			\midrule
			\multirow{6}{*}{Example} &  \imgitem{figures/user_e1.png}
			&  \imgitem{figures/user_e2.jpg}
			&  \imgitem{figures/user_e3.jpg}
			&  \imgitem{figures/user_e4.jpg}
			&  \imgitem{figures/user_e5.jpg} \\ 
			& & & & \\ 
			& & & & \\
			& & & & \\
			& & & & \\ 
			& & & & \\ \midrule
			& \multicolumn{5}{c}{\textbf{Visual ChatGPT}~\cite{wu2023visual}} \\
			Prompt length &  7.0 & 9.3 & 9.8 & 7.8 & 9.4 \\
			\# Try & 1.2 & 4.8 & 9.3 & 8.5 & 10 \\
			Score (1-5) & 4.2 & 2.2 & 0.7 & 0.2 & 0.0 \\
			\multirow{6}{*}{Example Result} &  \imgitem{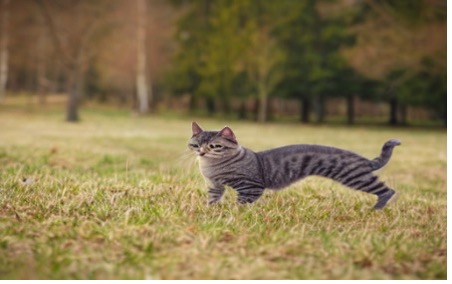}
			&  \imgitem{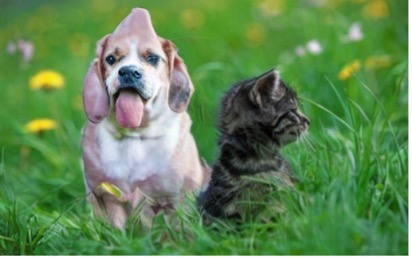}
			&  \imgitem{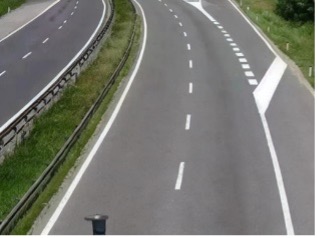}
			&  \imgitem{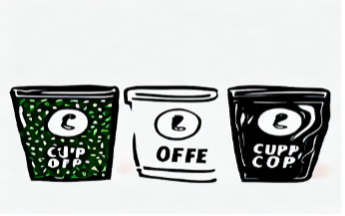}
			&  \imgitem{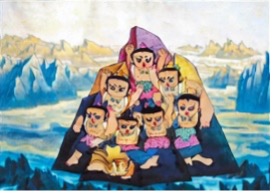} \\ 
			& & & & \\ 
			& & & & \\
			& & & & \\
			& & & & \\ 
			& & & & \\ \midrule
			& \multicolumn{5}{c}{\textbf{InternGPT (ours)}} \\
			Prompt length &  6.6 & 7.0 & 6.1 & 6.9 & 7.3 \\
			\# Try & 1.0 & 1.0 & 1.2 & 1.0 & 4.6 \\
			Score (1-5) & 4.5 & 5.0 & 3.8 & 4.4 & 1.9 \\
			\multirow{6}{*}{Example Result} &  \imgitem{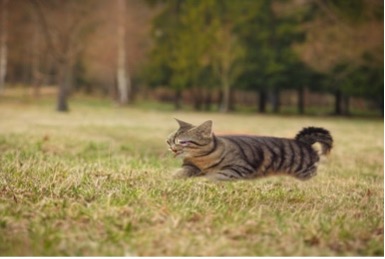}
			&  \imgitem{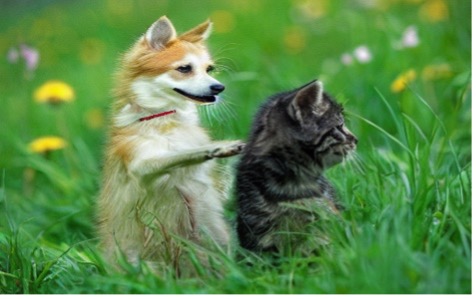}
			&  \imgitem{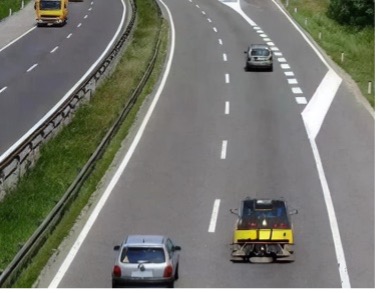}
			&  \imgitem{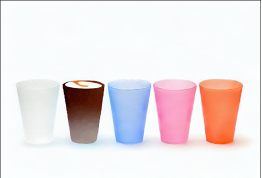}
			&  \imgitem{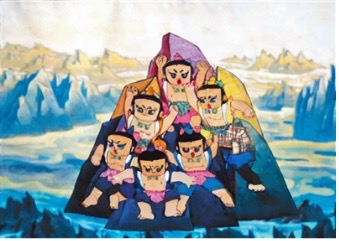} \\ 
			& & & & \\ 
			& & & & \\
			& & & & \\
			& & & & \\ 
			& & & & \\ \bottomrule
		\end{tabular}
		\vspace{1em}
		\caption{\textbf{User study for ``\texttt{replace something with something}''} ``\# Try'' indicates the number of attempts to get a satisfactory result, not more than 10 times. ``-'' means trying more than 10 times and still getting bad results.}
		\label{tab:user_study2}
	\end{table}
	
\begin{figure*}[t]
		\begin{minipage}{0.5\textwidth}
			\centering
			\includegraphics[width=1.0\linewidth]
			{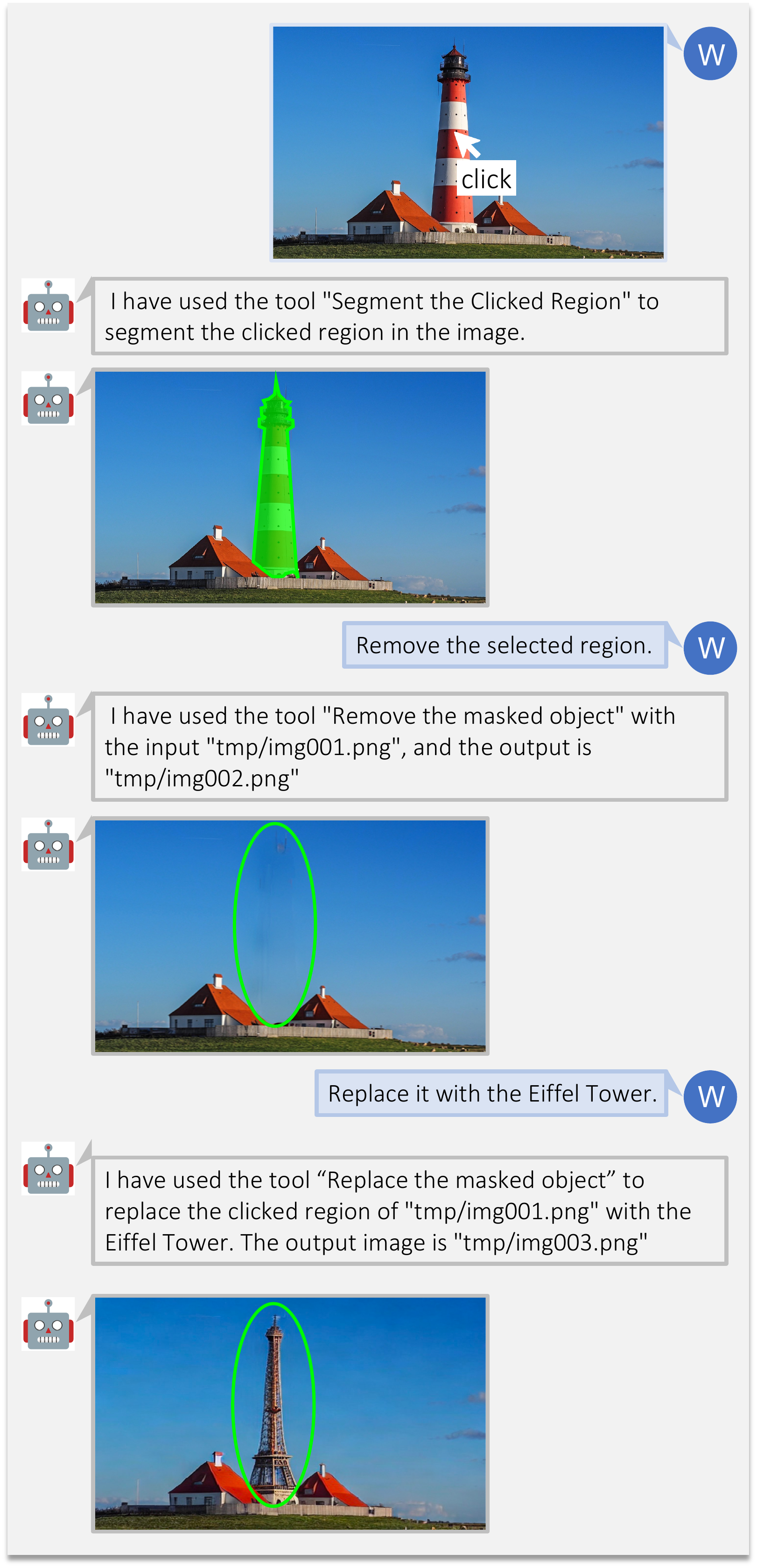}   
			\caption{
				\textbf{Interactive image editing.}
			}\label{fig:demo1}
		\end{minipage} \hfill
		\begin{minipage}{0.5\textwidth}
			\centering
			\includegraphics[width=1.0\linewidth]
			{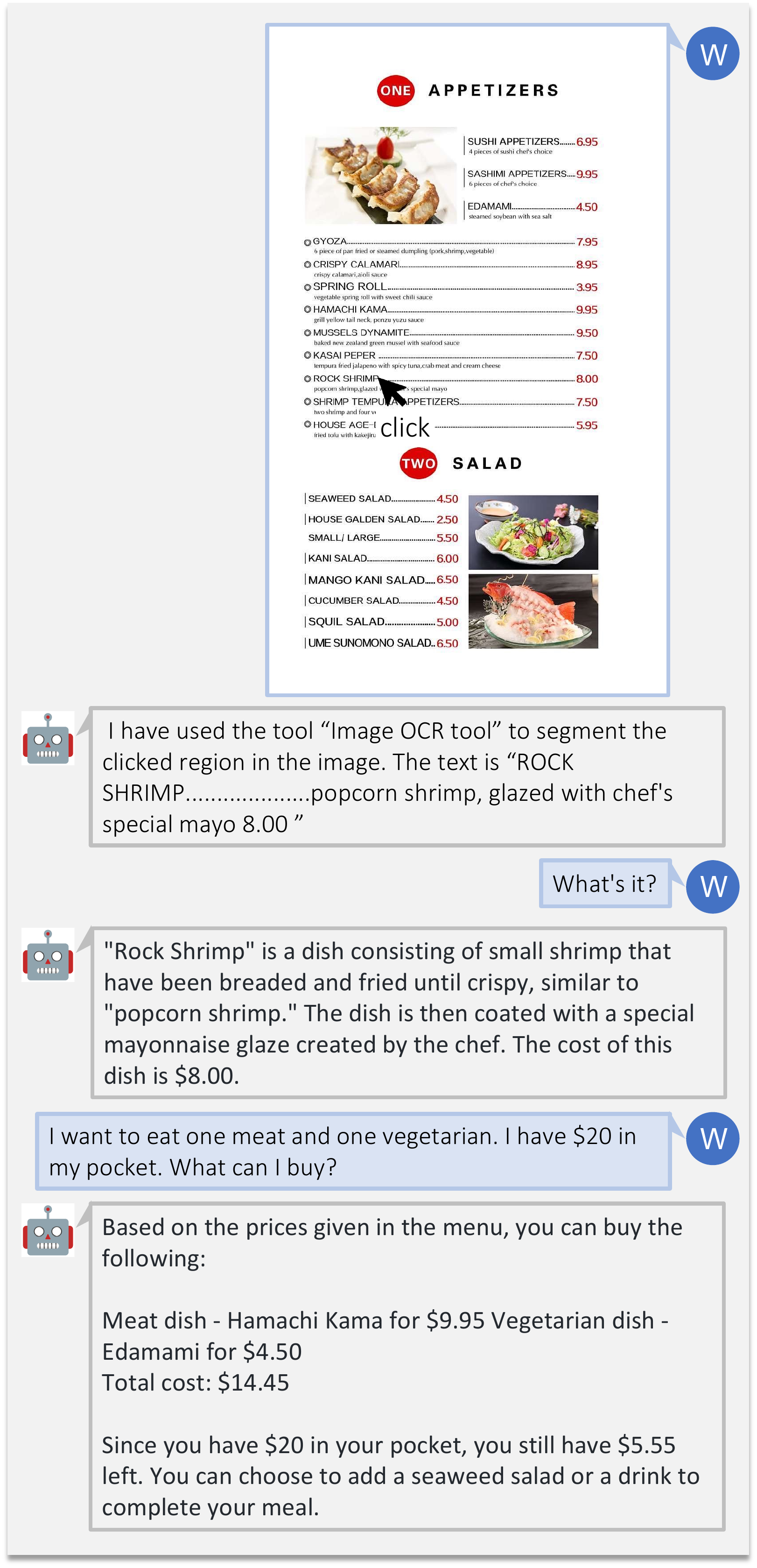}   
			\caption{
				\textbf{Interactive visual question answer.}
			}\label{fig:demo2}
		\end{minipage} \hfill
	\end{figure*}

	\subsection{Demonstrations}
	
	iGPT is a versatile framework that can support a range of applications involving verbal and non-verbal interactions. Users can communicate with the system using natural language, as well as gestures such as clicking, dragging, pointing, etc.
	We showcase several examples of interesting vision-centric applications:
	
	\textbf{Demo 1: interactive image editing.}
	Figure~\ref{fig:demo1} exhibits interactive image editing examples. Except for verbal instructions and an uploaded image, the iGPT receives a cursor clicking at the desired operating location of the image. The cursors interaction supplements user instructions to the LLM which controls the visual perception and generation instruments. With more concise and precise instructions, the InternGPT demonstrates pleasing image editing performance.
	
	\textbf{Demo 2: interactive visual question answering.}
	Figure \ref{fig:demo2} presents interactive visual question-answering examples. In the image-centric conversation with an intellective chatbot, the user's instructions may revolve around just a region of the image, rather than the entire image. The user can indicate the location through the simplest clicking or touching operation and communicate with the chatbot about the concerned region.
	
\begin{figure*}[t]
		\begin{minipage}{0.5\textwidth}
			\centering
			\includegraphics[width=0.9\linewidth]
			{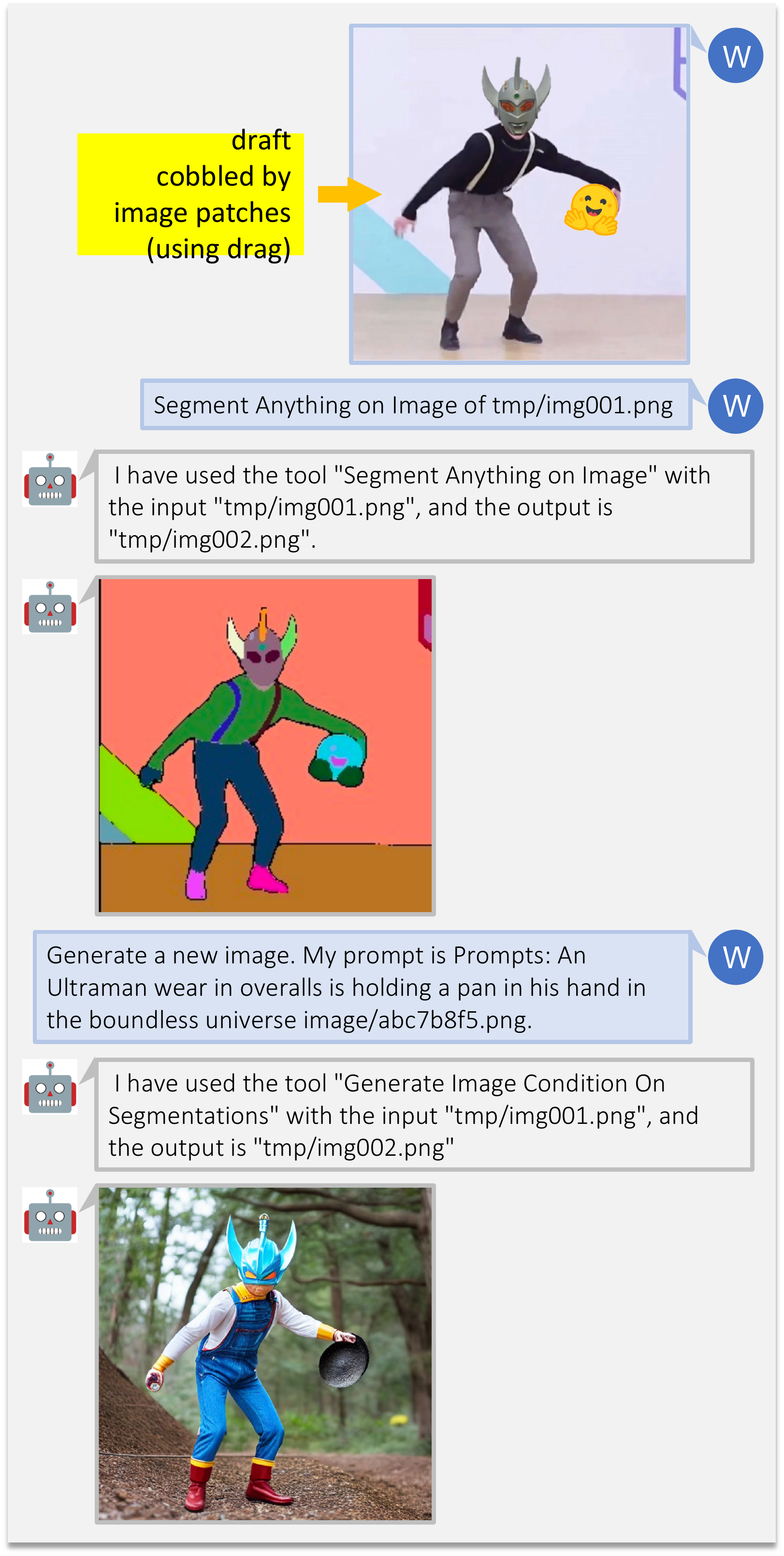}   
			\caption{
				\textbf{Interactive image generation.}
			}\label{fig:demo3}
		\end{minipage} \hfill
		\begin{minipage}{0.5\textwidth}
			\centering
			\includegraphics[width=0.9\linewidth]
			{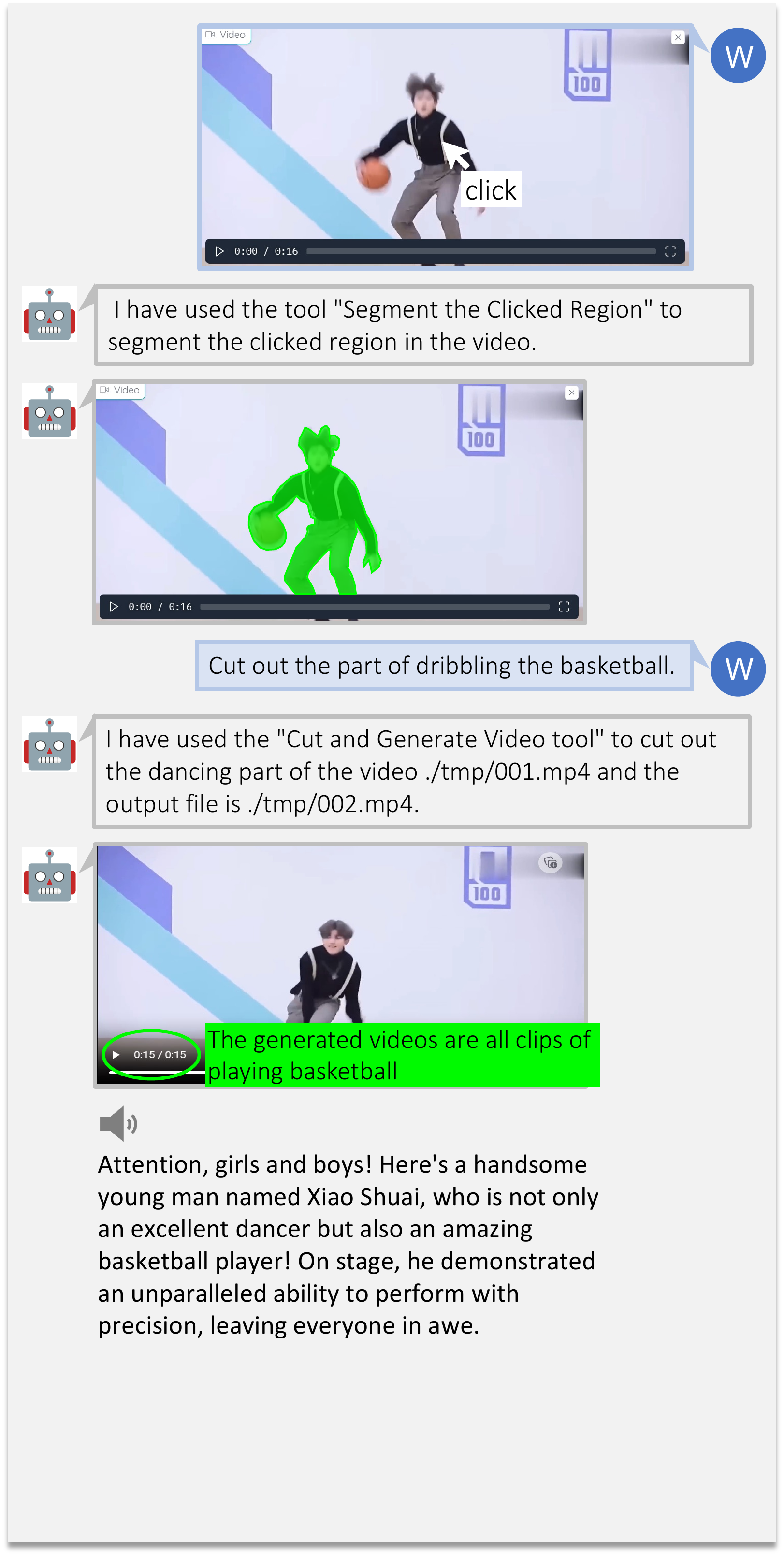}   
			\caption{\textbf{Video highlight interpretation.}}\label{fig:demo4}
		\end{minipage} 
	\end{figure*}

	\textbf{Demo 3: interactive image generation.}
	Figure \ref{fig:demo3} demonstrates image creation examples. The iGPT stores multiple picking image parts as materials. The user can easily assemble the materials by the dragging interaction. After receiving a verbal prompt and the assembled images from the user, the iGPT controls the generation instruments to create a fancy image, and give it a title.
	
	\textbf{Demo 4: video highlight interpretation.}
	We have implemented a feature called Video Highlight Interpretation, which is demonstrated in Figure~\ref{fig:demo4}. This feature is designed to help users quickly and easily understand the essence of a video without having to watch the entire thing. iGPT locates the beginning and end of a video highlight based on its temporal position and then provides a lighthearted and humorous voice-over interpretation of the highlighted content. In addition to its primary use, this feature can also be leveraged by users who want to quickly create derivative works based on a popular video's main points. By using this feature, users can save time and effort in creating new content based on existing videos, while still ensuring that the new content accurately reflects the original material.
	
	\vspace{-3pt}

	These examples exhibit the convenient interactive capabilities and powerful performance on vision-centric tasks, which come from the novel design of the intelligence system.
	
	\subsection{Demonstrations of Husky} 
\begin{figure}[htp]
		\centering
		\includegraphics[width=1.05\linewidth]{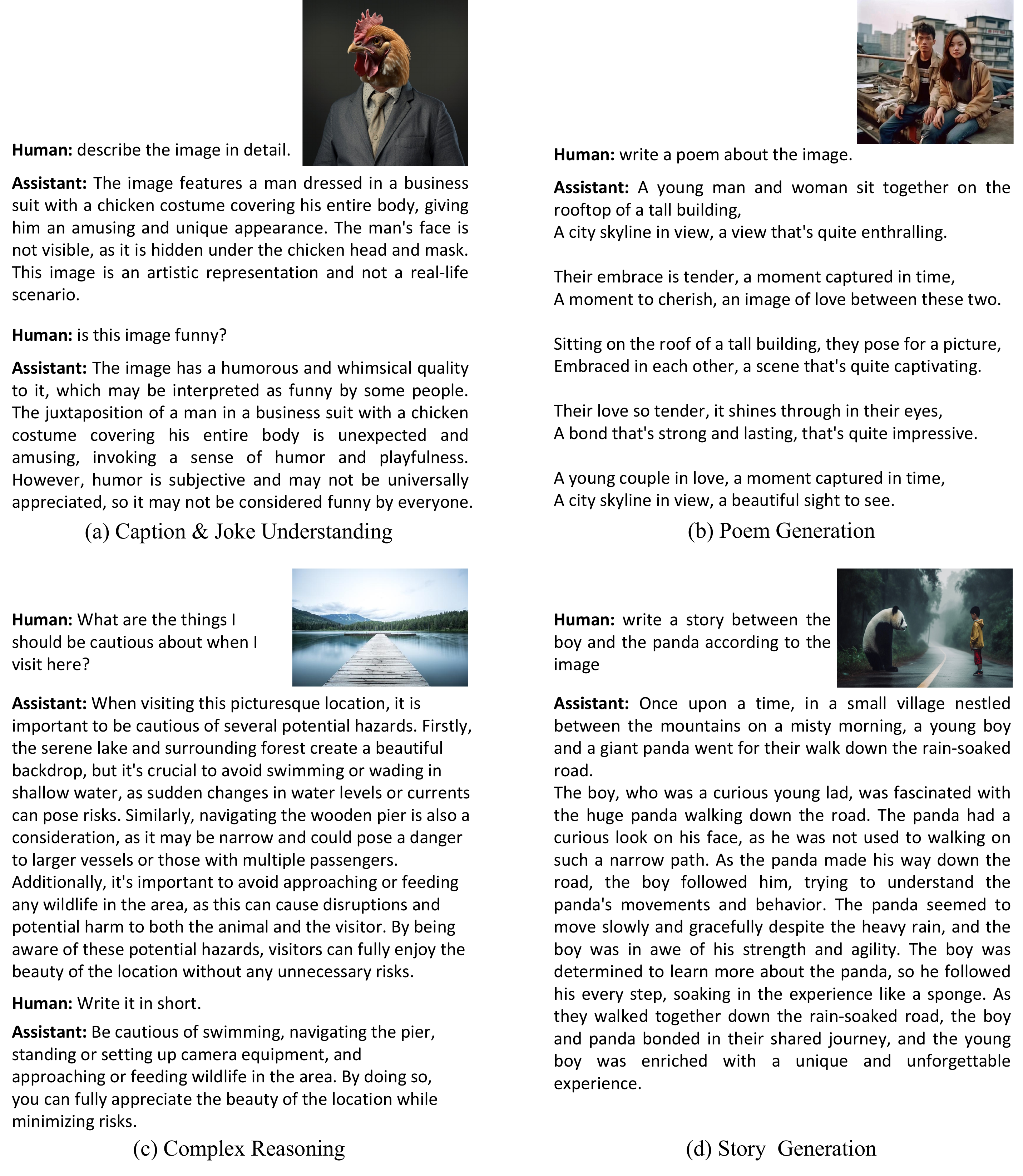}
		\caption{\textbf{Demonstrations of Husky across various scenarios.} }
		\label{fig:husky_6}
	\end{figure}

	\textbf{Qualitive Evaluation. }
	In this section, we present demonstrations across various scenarios to assess the performance of the large vision-language model Husky. Qualitive results are shown in Figure~\ref{fig:husky_6}. These diverse examples showcase the strong capabilities of our Husky.

	\textbf{Quantitative Evaluation. }
	We also perform a quantitative evaluation on Husky. We follow the setting of LLaVA~\cite{liu2023llava} and adopt Husky to predict the answers to the provided 90 questions of the 30 COCO validation images. We leverage ChatGPT-3.5-turbo to measure the quality of our model’s generated responses and apply GPT-4's predictions (provided by LLaVA) as a reference. As reported in Table~\ref{tab:quantitative_results}, Husky can impress ChatGPT-3.5-turbo with 93.89 \% GPT-4 Quality. 
	It is worth mentioning that, according to the analysis conducted by ChatGPT-3.5-turbo, Husky demonstrates a reasoning capacity that is comparable to that of GPT-4. However, when it comes to conversation and providing detailed descriptions, Husky is still lagging behind that of GPT-4.

	\begin{table*}[htp]
		\begin{center}
			\begin{tabular}{ccccc}
				\toprule[1.2pt]
				Conversation & Detail description & Complex reasoning & All\\ 
				\midrule[1.2pt]
				96.13 & 83.87 & 102.95 & 93.89 \\
				\bottomrule[1.2pt]
			\end{tabular}
		\end{center}
		\caption{Detailed Quantitative Results}
		\label{tab:quantitative_results}
	\end{table*}

	\subsection{Limitations}
	Given that iGPT is built upon online resources, potential limitations of this integrated system may include:
	\textbf{Model Performance.} iGPT's effectiveness largely hinges on the quality and accuracy of the underlying open-sourced models. Limitations or biases in these models could adversely affect iGPT's performance.
	
	\textbf{Scalability.} As user interactions grow more complex or encompass a greater number of instances, maintaining accuracy and response times could prove challenging for the system. Moreover, the current non-learnable cooperation between vision foundation models and language models, e.g., not being tuned by instruction data, could impede capitalizing on the full capacity of the used models.
	
	\textbf{Adaptability.} iGPT might struggle to adjust to novel or uncommon scenarios absent from the training data of its employed models, leading to a compromised performance in unpredictable situations.
	
	\textbf{User Interface.} Despite emphasizing user-friendliness, some users might still face difficulties utilizing the combined pointing and language instructions effectively, which could impact their overall experience.
	
	\textbf{Compatibility.} Achieving seamless integration with a diverse array of devices and platforms could pose challenges due to varying hardware capabilities, software constraints, or accessibility requirements.
	
	\textbf{Privacy and Security.} As an AI-driven system, iGPT may raise concerns regarding data privacy and security, especially if sensitive information is processed or shared within the platform.

	\section{Conclusion}
	We have developed InternGPT (iGPT for short), a dynamic visual framework that tackles vision-centric tasks. Designed with an emphasis on user-friendliness and efficacy, iGPT delivers a top-tier experience among current open-source tools. Our methodology seamlessly merges pointing and text commands to issue instructions, allowing users to harness the power of various open-sourced models without needing expert knowledge. The robustness of iGPT has been showcased in complex visual scenarios involving multiple instances, utilizing user-level text comprehension from the current LLM and employing cursors or gestures through pointing devices.
	
	iGPT appreciates both pointing and linguistic directives, leveraging the perception unit and LLM controller to synchronize and execute applications within an open-world toolkit. Our system has successfully performed intricate interactive tasks beyond the capabilities of purely language-based systems. User surveys have demonstrated that integrating pointing and language instructions can boost work efficiency in challenging situations. Aspiring to be the foundational benchmark for visual interactive systems, iGPT is committed to ongoing updates and improvements for exceptional performance.
	
	\clearpage
	
	{
		\small
		\bibliographystyle{plain}
		\bibliography{egbib}
	}
	
\end{document}